\documentclass{article}

\usepackage{arxiv}

\usepackage[utf8]{inputenc} 
\usepackage[T1]{fontenc}    
\usepackage{hyperref}       
\usepackage{url}            
\usepackage{booktabs}       
\usepackage{amsfonts}       
\usepackage{nicefrac}       
\usepackage{microtype}      
\usepackage{lipsum}		
\usepackage{graphicx}
\usepackage{doi}
\usepackage{xcolor}
\usepackage{multirow}
\usepackage{siunitx}

\renewcommand{\undertitle}{}
\renewcommand{\shorttitle}{Generative Language Models with RAG for Automated Short Answer Scoring}

\title{Generative Language Models with Retrieval Augmented Generation for Automated Short Answer Scoring}

\author{\href{https://orcid.org/0000-0002-8961-4302}{\includegraphics[scale=0.06]{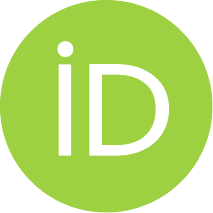}\hspace{1mm}Zifan Wang\thanks{This work was completed during an internship at Cambium Assessment.} } \\
	Cambium Assessment\\Syracuse University\\
	\texttt{zwang345@syr.edu} \\
	\And
	\href{https://orcid.org/0000-0002-2129-7021}{\includegraphics[scale=0.06]{orcid.pdf}\hspace{1mm}Christopher Ormerod} \\
	Cambium Assessment\\
	\texttt{christopher.ormerod@cambiumassessment.com} \\
}

\date{}

\begin{document}
\newcommand{\todo}[1]{\textcolor{red}{#1}}

\maketitle

\begin{abstract}
Automated Short Answer Scoring (ASAS) is a critical component in educational assessment. While traditional ASAS systems relied on rule-based algorithms or complex deep learning methods, recent advancements in Generative Language Models (GLMs) offer new opportunities for improvement. This study explores the application of GLMs to ASAS, leveraging their off-the-shelf capabilities and performance in various domains. We propose a novel pipeline that combines vector databases, transformer-based encoders, and GLMs to enhance short answer scoring accuracy. Our approach stores training responses in a vector database, retrieves semantically similar responses during inference, and employs a GLM to analyze these responses and determine appropriate scores. We further optimize the system through fine-tuned retrieval processes and prompt engineering. Evaluation on the SemEval 2013 dataset demonstrates a significant improvement on the SCIENTSBANK 3-way and 2-way tasks compared to existing methods, highlighting the potential of GLMs in advancing ASAS technology.
\end{abstract}


\section{Introduction}

An effective assessment program employs various question formats, each designed to evaluate specific standards. Standards involving comprehension and knowledge are often best assessed using short answer questions, requiring students to construct their own responses. Automated Short Answer Scoring (ASAS) uses natural language processing (NLP) and statistical models to evaluate student responses to open-ended questions. The earliest statistical models were frequency and rule-based model \cite{leacock_c-rater_2003}. As methods in NLP have advanced, the models applied to ASAS have become more sophisticated. These methods include clustering methods \cite{basu_powergrading_2013}, a mixture of recurrent and feed-forward neural networks \cite{dong_attention-based_2017}, language models \cite{sung_pre-training_2019}, and ensembles of networks \cite{ormerod_automated_2022}. 

Creating an Automated Short Answer Scoring (ASAS) system is primarily costly due to the need for high-quality, manually scored responses for training. A significant challenge in ASAS development is reducing the required number of training samples while maintaining scoring quality. Recent advancements in generative language models (GLMs) have led to promising developments in zero-shot, one-shot, and few-shot learning techniques \cite{wang_generalizing_2020}. Applying these approaches could significantly reduce the number of training responses needed for a high-quality scoring pipeline. Our research aims to leverage these methods, utilizing advanced GLMs like ChatGPT \cite{openai_gpt-4_2023} and Claude \cite{anthropic2024claude}, to maximize scoring quality with minimal training samples.

In this study, we focus on the SemEval-2013 joint student response dataset \cite{dzikovska_semeval-2013_2013}. This dataset has been subjected to many traditional approaches~\cite{ott2013comet, heilman2013ets, jimenez2013softcardinality, sultan2016fast, ramachandran2015generating, saha2018sentence, marvaniya2018creating, sung2019improving}. Pretrained encoder-based language models like BERT \cite{devlin_bert_2018} and ELECTRA \cite{clark_electra_2020} have set impressive benchmarks \cite{sung_pre-training_2019}.  

Large proprietary models excel in zero-shot and few-shot learning, their performance in ASAS still falls short of production-level systems \cite{williamson_framework_2012}. To bridge this gap, we employ two strategies that leverage the training corpus:
\begin{enumerate}
    \item We implement a pipeline of a retrieval-augmented generation (RAG) and Generative Language Model (GLM) for Automated Short Answer Scoring (ASAS).
    \item We optimize the scoring performance by fine-tuning the information retrieval (IR) component \cite{zamani2022retrieval}.
    \item We optimize prompt templates for GLMs in ASAS by Claude Prompt Generator~\cite{anthropic2023claude} and dynamically choose templates for different scenarios. 
\end{enumerate}

This paper is organized as follows: We introduce background about datasets, information retrieval, generative language models, retrieval augmented generation, and prompt optimization in Section~\ref{sec:background}. We describe the particular way in which we fine-tune our IR pipeline, our implementation of RAG, and the manner in which we optimize prompts in Section~\ref{sec:method}. We present our experiments, which includes metrics used, and the relevant results in Section~\ref{sec:experiments}. Lastly, we conclude with a discussion of future research directions in Section~\ref{sec:discussion}. 
\section{Background}\label{sec:background}

\subsection{Dataset}

For this study, we utilized the dataset from the SemEval-2013 task 7 on joint student responses~\cite{dzikovska_semeval-2013_2013}. This dataset contains two distinct corpora; the BEETLE dataset and SciEntsBank dataset. The BEETLE dataset consists of 3000 student responses to a set of 56 questions relating to electricity and electrons. The SciEntsBank consists of approximately 10,000 answers to a range of 197 questions relating to 15 different scientific domains. Most responses were 1 or 2 sentences long and assessed specific comprehension skills relating to the material provided. This serves as an interesting use case for the method we propose because traditional ASAS systems require a single model per question, however, each corpus contains relatively few answers per question.

The dataset underwent manual categorization using three distinct classification methods: a 5-category system, a 3-category system, and a 2-category system. In the 5-category classification, each student's answer was assigned to one of the following groups:
\begin{itemize}
\item {\bf Correct:} A paraphrase of the reference answer.
\item {\bf Partially correct incomplete:} Contains some but not all information from the reference answer.
\item {\bf Contradictory:} The student answer and reference answer contradict each other.
\item {\bf Irrelevant:} Provides information on the topic but not relevant to the answer.
\item {\bf Non-domain:} Does not provide domain content. e.g., ``I don't know"
\end{itemize}
The 3-way classification was derived from the 5-way scheme by grouping {\bf Irrelevant}, and {\bf Non-domain} responses into a single {\bf Incorrect} category. The 2-way simplifies the 5-way scheme by categorizing all responses except those labeled {\bf Correct} as {\bf Incorrect}.

Both the BEETLE dataset and the SciEntsBank dataset have two distinct test sets according to the conditions of the test:
\begin{itemize}
\item {\bf Unseen answers (UA):} A portion of student responses for each question in the training dataset was randomly selected and set aside. This reserved subset forms a separate evaluation set, allowing assessment of the system's performance on familiar questions (those present in the training data) but with previously unseen student answers.
\item {\bf Unseen questions (UQ):} An evaluation set was created to measure the system's effectiveness on new questions within the same subject areas as the training data. This set consists of all student responses to a randomly chosen subset of questions from each dataset, which were entirely withheld from the training process.
\end{itemize}
In addition to these two test sets, the SciEntsBank dataset contains a third test set:
\begin{itemize}
\item {\bf Unseen Domain (UD):} An evaluation set composed of responses covering domains absent from the training data, designed to assess the system's performance across unfamiliar domains.
\end{itemize}

\begin{table}[htb]
\caption{Distribution of labels in BEETLE and SciEntsBank datasets.}
\centering
\begin{tabular}{l|c|cc|c|c|ccc c} \toprule
\multirow{2}{*}{label} & \multicolumn{4}{c|}{BEETLE} & \multicolumn{5}{c}{SCIENTSBANK} \\
 & train (\%) & UA & UQ & Test-Total (\%) & train (\%) & UA & UQ & UD & Test-Total (\%) \\ \midrule
correct & 1665 (42\%) & 176 & 344 & 520 (0.41\%) & 2008 (40\%) & 233 & 301 & 1917 & 2451 (42\%) \\
pc inc & 919 (23\%) & 112 & 172 & 284 (23\%) & 1324 (27\%) & 113 & 175 & 986 & 1274 (22\%) \\
contra & 1049 (27\%) & 111 & 244 & 355 (28\%) & 499 (10\%) & 58 & 64 & 417 & 539 (9\%) \\
irrlvnt & 113 (3\%) & 17 & 19 & 36 (3\%) & 1115 (22\%) & 133 & 193 & 1222 & 1548 (27\%) \\
non dom & 195 (5\%) & 23 & 40 & 63 (5\%) & 23 (0.5\%) & 3 & 0 & 20 & 23 (0.4\%) \\ \bottomrule
\end{tabular}

\label{table:label_distribution}
\end{table}

Table \ref{table:label_distribution} contains a summary of the 5-way distributions from which one may derive the 3-way and 2-way distributions. What differentiates this task from typical ASAS is that the 5-way and 3-way classification schemes are not ordinal, which is naturally the consequence of a holistic rubric. This distinguishes the task as a classification task from the Kaggle ASAS dataset \cite{shermis_state---art_2014} and the Powergrading dataset \cite{basu_powergrading_2013}. Traditional metrics for evaluating the performance of automated scoring systems \cite{williamson_framework_2012}, like Cohen's quadratic weighted kappa, do not apply. This also means traditional regression-based approaches that deliver impressive performance for holistic rubrics, such as the Kaggle dataset (e.g., \cite{yang_enhancing_2020, ormerod_mapping_2022}), should not be applied. 

\subsection{Information Retrieval}

\begin{figure*}[!b]
    \centering
  \includegraphics[width=0.85\columnwidth]{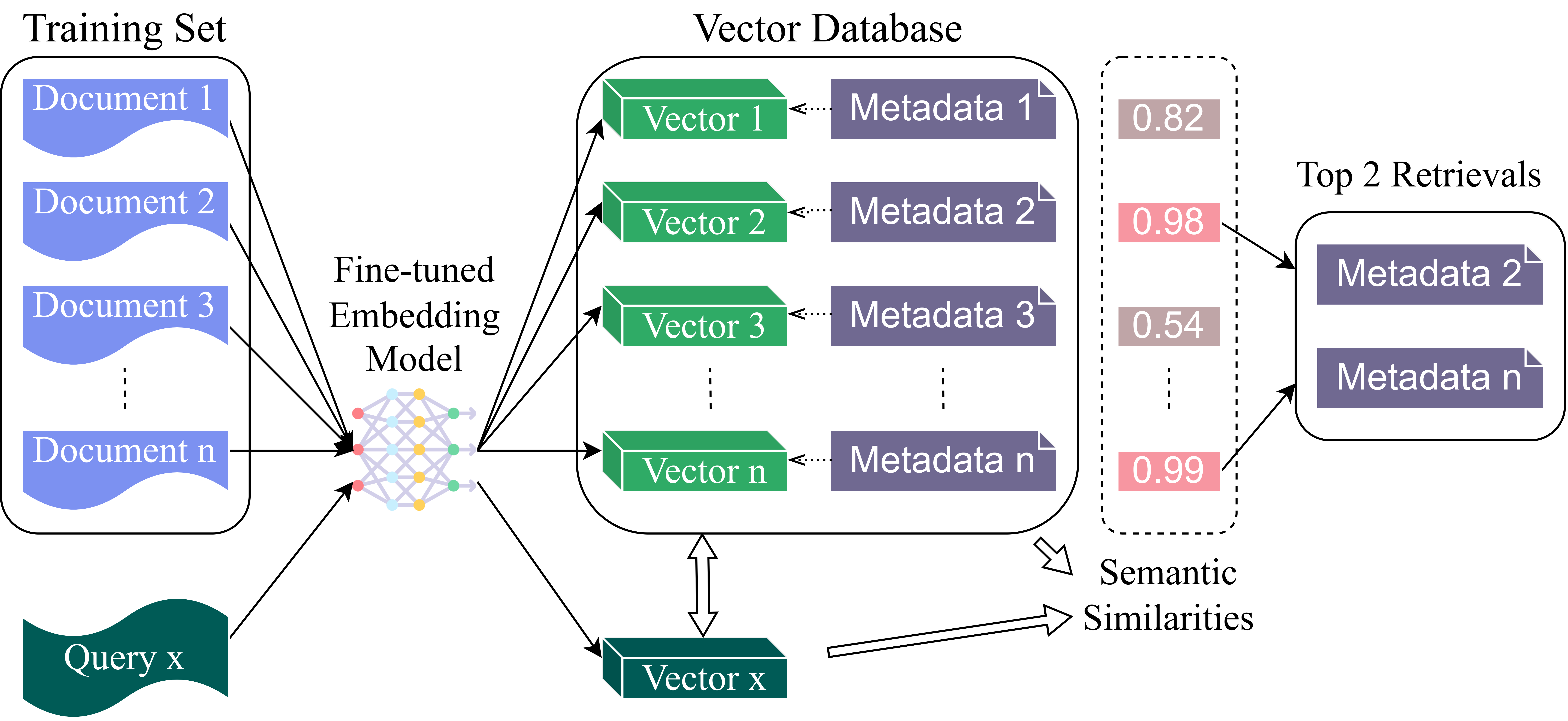}
  \caption{Information Retrieval Workflow}\label{fig:ir}
\end{figure*}

An Information Retrieval (IR) pipeline is a system that provides relevant information (i.e., documents) from a database based on user queries. The key to this approach is to map a document to a vector that encodes information relevant to queries, called its representation, in some vector space. The idea is that semantically similar documents are mapped to similar vectors. More formally, if $d_1$ and $d_2$ are documents, and $\phi$ is an embedding that maps the set of documents to $\mathbb{R}^n$, then the semantic similarity between the documents, $S(d_1,d_2)$, is modeled by the cosine similarity of the embeddings, given by   
\begin{equation}\label{equ:cosine_sim}
S(d_1, d_2) = \dfrac{\phi(d_1) \cdot \phi(d_2)}{\left\| \phi(d_1) \right\| \left\| \phi(d_2) \right\|}.
\end{equation}
Consider a set of documents, $D$, consisting of elements $d_1, \ldots, d_n$. We can represent their normalized embeddings as rows in a matrix $A$. If we can convert a query into a vector $x$, then finding the $k$ most relevant documents to that query in $D$ becomes equivalent to identifying the $k$ largest values in the vector resulting from the matrix multiplication $A x$. This process is illustrated in Figure \ref{fig:ir}.

In our case, $D$ is a set of student responses In the training set, our query is a response, $d$, in the test set. Since we expect $d$ to be of the same form as the elements of $D$, it makes sense that $x = \phi(d)$. In settings like question-answering systems, where the queries are of a different form to the documents being retrieved, the two embeddings can be different \cite{reimers2019sentence}. Examples of more traditional methods of determining the embedding, $\phi$, have included term frequency-based approaches like {\em tf}-{\em idf} \cite{singhal2001modern} and recurrent approaches \cite{chen2017reading}. These approaches have been superseded by transformer-based approaches \cite{reimers2019sentence}.

Many researchers use models pretrained on large datasets of paraphrase pairs, like the Quora Question Pairs. Our observation is that the SemEval-2013 dataset contains responses classified as paraphrases of a reference answer (labeled as "correct"). Since paraphrasing is transitive, all "correct" responses to a given question are effectively paraphrases of each other. Additionally, any response not labeled as "correct" is, by definition, not a valid paraphrase of the reference answer. This structure gives us a valuable set of positive and negative examples. We can use these to fine-tune a pretrained model, allowing it to make more nuanced distinctions between classes of responses that are specific to our dataset's characteristics and requirements. We use this model to construct a vector database of training responses.

\subsection{Generative Language Models}

A Generative Language Model (GLM) is a type of neural network trained on massive amounts of text data based on deep learning techniques. The most ubiquitous examples in the literature have become transformer-based models \cite{vaswani_attention_2017}. The most general architecture consists of an encoder, which maps text to a vector space, and a decoder that maps that vector space back to text. The advantage these networks have is that we can improve the performance on downstream tasks by pertaining these models on large corpora of data \cite{devlin_bert_2018, radford_improving_2018}. There are three classes of such models; encoder networks such as BERT \cite{devlin_bert_2018}, decoder networks such as GPT \cite{radford_improving_2018}, and encoder-decoder networks such as T5 \cite{raffel_exploring_2020}. 

The embedding we use as part of our IR pipeline is a pretrained encoder network trained for semantic similarity \cite{reimers2019sentence} derived from a truncation of the BERT model \cite{devlin_bert_2018}. We fine-tune this model for our own use case. 

The second type of model we use is a decoder network. The fundamental task of decoder GLMs is to predict the next token from a sequence of prior tokens \cite{radford_improving_2018}. Modern GLMs generally have an order of magnitude more parameters and are trained in several stages. The first stage is to pretrain the model on a large corpus of data. The model is then fine-tuned to perform instructions \cite{radford_improving_2018}. Lastly, the model is subjected to reinforcement learning \cite{ouyang2022training}. The remarkable thing about these models is that they are often capable of performing tasks not necessarily similar to tasks in their training data \cite{anthropic2024claude, openai_gpt-4_2023}.

\subsection{Retreival Augmented Generation}

Even though GLMs possess extensive knowledge and demonstrate excellent performance on various tasks, including auto-scoring, the vanilla auto-scoring approach—providing a GLM with answers, questions, and rubrics, and asking it to give feedback—presents challenges. Grading answers solely based on questions and rubrics is difficult, even for GLMs, let alone that rubrics are not always readily available. 

Studies indicate that providing additional context and examples can enhance their performance. Therefore, we incorporate documents retrieved by our IR system as examples in the prompt for the GLM to reference. This method enables GLMs to learn from the examples and improve their grading capabilities, with or without a rubric, by implicitly conveying the logic of grading through the examples.

\subsection{Prompt Optimization}
Recent studies have demonstrated that carefully crafted prompts can significantly enhance performance across various tasks. Effective prompts typically employ several key techniques:

\begin{itemize}
    \item Structured format: Organize the prompt to clearly present descriptions, requirements, examples, procedures, and scoring criteria.
    \item Role assignment: Explicitly define the role the Generative Language Model should assume.
    \item XML tag utilization: Employ XML tags to delineate input and output fields, enhancing clarity and structure.
    \item Task-specific techniques: Incorporate methods like chain-of-thought reasoning when appropriate, though this may not be suitable for all applications, such as autoscoring.
\end{itemize}

While we initially drafted our prompts incorporating these techniques, we recognized the need for further optimization. To address this, we explored two advanced tools: DSPy~\cite{khattab2023dspy, khattab2022demonstrate} and the Claude Prompt Generator. These tools offer distinct approaches to prompt refinement, each with its own strengths and limitations.

\subsubsection{DSPy}
DSPy, described as "a framework for algorithmically optimizing LM prompts and weights" \cite{khattab2023dspy}, provides a systematic approach to prompt optimization. Our focus was primarily on its prompt optimization capabilities. The core concept behind DSPy involves the interaction between two GLMs:
\begin{itemize}
    \item A low-temperature task GLM: Responsible for evaluating prompt performance with high consistency.
    \item A high-temperature prompt critic GLM: Tasked with improving prompts based on optimization history, leveraging its high temperature setting to generate creative alternatives.
\end{itemize}

The optimization process in DSPy follows these steps:

\begin{enumerate}
    \item Initialization: Input a dataset, evaluation metric, and draft prompt (including role setting and input/output field descriptions).
    \item Iterative optimization: Conduct $D$ optimization steps.
    \item Candidate generation: For each step, generate $B$ prompt candidates.
    \item Evaluation: Assess all candidates on the entire dataset.
    \item Ranking and selection: Rank all candidates (including previous ones) by performance, retaining only the top $B$ candidates.
\end{enumerate}

This iterative optimization process enables DSPy to explore a wide range of potential prompts, ultimately yielding an optimized version through a process of guided randomization and performance-based selection.

\subsubsection{Claude Prompt Generator}

The Claude Prompt Generator~\cite{anthropic2023claude}, an integrated tool within the Claude Console, offers a different approach to prompt optimization. This tool leverages the Claude3.5 Sonnet model to refine user-provided prompts based on a comprehensive, expert-calibrated meta-prompt. The meta-prompt incorporates multiple examples and guidelines, allowing the model to draw upon its extensive knowledge and understanding to enhance the initial prompt.

\subsubsection{Comparative Analysis of DSPy and Claude Prompt Generator}

Both tools offer unique advantages and face certain limitations in the context of prompt optimization.

\textbf{DSPy:}
\begin{itemize}
    \item Strengths:
    \begin{itemize}
        \item Automated search capability: Systematically explores a wide range of prompt variations.
        \item Performance-based selection: Retains and builds upon the most effective prompts.
        \item Potential for discovering novel prompt structures: The high-temperature critic GLM can generate creative alternatives.
    \end{itemize}
    \item Limitations:
    \begin{itemize}
        \item Rigid template structure: Relies on a fixed template with predefined sections, potentially limiting flexibility.
        \item Scope of optimization: The prompt critic GLM can only modify the system prompt and output field names, leaving other structural elements unchanged.
        \item Potential outdated practices: The fixed template may not incorporate the latest advancements in prompt engineering.
    \end{itemize}
\end{itemize}

\textbf{Claude Prompt Generator:}
\begin{itemize}
    \item Strengths:
    \begin{itemize}
        \item Up-to-date meta-prompt: Incorporates recent best practices and well-crafted examples.
        \item Efficiency: Provides refined prompts quickly without requiring multiple iterations.
    \end{itemize}
    \item Limitations:
    \begin{itemize}
        \item Lack of iterative refinement: Does not incorporate performance feedback or allow for multiple optimization cycles.
        \item Complexity of evaluation: Assessing the performance of generated prompts is challenging and not directly integrated into the tool.
    \end{itemize}
\end{itemize}

In our experimentation, we found that the Claude Prompt Generator often produced superior results compared to DSPy, likely due to its more current meta-prompt and holistic approach to prompt refinement. However, the lack of iterative optimization in the Claude Prompt Generator presents a potential area for future improvement, particularly in scenarios where prompt performance can be quantitatively measured.

\section{Method}\label{sec:method}

The proposed autoscoring system comprises two main phases: an offline optimization phase and an online running phase. The offline phase focuses on preparing datasets, building the Information Retrieval (IR) system, and generating optimal prompts, while the online phase utilizes these components for real-time autoscoring.

\subsection{Offline Optimization Phase}

The offline optimization phase consists of three critical steps: training the IR system, constructing the IR vector database, and optimizing GLM prompt templates.

\subsubsection{Training the IR System}

\textbf{Dataset Preparation:} A comprehensive dataset for autoscoring typically includes questions, rubrics, reference answers, student responses, and corresponding judgments. However, due to practical constraints such as cost and policy restrictions, we prioritize the essential components: student responses and their associated judgments. The dataset is divided into training and test sets, with the latter held out during the training process to ensure unbiased evaluation.

We follow 4 steps to prepare a training set for every dataset to train the Information Retrieval (IR) system:
\begin{enumerate}
    \item Division by Question: We first divide the training set into sub-training sets, each corresponding to a specific question.
    \item Answer Pair Formation: Within each sub-training set, we iterate through every student answer, pairing it with all other answers to form answer pairs. This process is done without repetition to avoid redundancy.
    \item Labeling Strategies: We implement and compare two strategies for labeling answer pairs:
   \begin{itemize}
       \item Strict Labeling: Pairs are labeled 1 only if both answers are from either the {\bf Correct} or {\bf Incorrect} categories. Pairs are labeled 0 if the answers are from different categories or from the same category but not {\bf Correct} or {\bf Incorrect}.
       \item General Labeling: Pairs are labeled 1 if both answers are from the same category, regardless of which category it is.
   \end{itemize}
   \item Balancing the Training Set: Both labeling strategies result in an imbalanced training set, with a higher proportion of 0-labeled pairs. To mitigate this imbalance and improve training efficacy, we create a balanced training set by retaining all pairs labeled 1 while randomly sampling an equal number of pairs labeled 0. This approach ensures that we include as many positively labeled pairs as possible while maintaining a balanced distribution between positive and negative examples, which is crucial for effective model training.
\end{enumerate}
   
For special training losses, such as triplet loss, we construct triplet training sets. Each triplet consists of (anchor answer, positive answer, negative answer). We build these sets by: 1) Iterating through each answer in the sub-training sets, designating it as the anchor answer, 2) Matching the anchor with an answer from the same category as the positive answer, 3) Matching the anchor with an answer from a different category as the negative answer.This process is done iteratively and without repetition to ensure comprehensive coverage of the answer space.

\textbf{Fine-tune IR Backbone.} We fine-tune a pre-trained Sentence Transformer model, 'all-MiniLM-L6-v2'~\footnote{The model checkpoint is hosted at \href{https://huggingface.co/sentence-transformers/all-MiniLM-L6-v2}{https://huggingface.co/sentence-transformers/all-MiniLM-L6-v2}. The model size is 80 MB.}, as the backbone embedding model in our Information Retrieval (IR) system, for a good balance between performance and computational efficiency.

To explore factors contributing to performance differences, we implement and compare two training strategies:

\begin{itemize}
    \item Question-specific Training: This approach involves training a separate IR backbone for each question. Each backbone is optimized specifically for its target question using the corresponding sub-training set. During the online running phase, the IR pipeline dynamically selects the appropriate backbone based on the question being evaluated. This strategy allows for highly specialized models but may require more computational resources and storage.
    
    \item Global Training: In this approach, we train a single global IR backbone for all questions in a dataset. The global training set is created by combining all question-specific sub-training sets. The training performance is optimized based on the overall loss across all questions. This strategy results in a more generalized model that can be applied to any question, potentially sacrificing some question-specific performance for improved efficiency and scalability.
\end{itemize}

To further optimize the IR backbone, we implement and compare three loss functions. For all three functions, the backbone training loss is averaged within every batch of inputs:

\begin{itemize}
    \item Cosine Similarity Loss: $loss=(cos(embed(answer_1), embed(answer_2)) - label)^2$, where $answer_1$ and $answer_2$ are the two answers in a pair, $label$ is the label of the pair, $embed(*)$ is the embedding operation using the backbone model, and $cos$ is presented in Equation \ref{equ:cosine_sim}.
    \item Cosine Sentence Loss: $loss=logsum(1+exp(cos(embed(i),embed(j))-cos(embed(k),embed(l)))+exp...)$, where $logsum(*)$  $(k,l)$ and $(i,j)$ are any of the answer pairs in the batch such that the expected similarity of $(k,l)$ is greater than $(i,j)$.
    \item Triplet Loss: $loss=max(|answer_a - answer_p| - |answer_a - answer_n| + margin, 0)$, where $answer_a$, $answer_p$ and $answer_n$ are anchor, positive, and negative answers, and $margin$ controls sensitivity. We set $margin$ as 3 in this work.
\end{itemize}

For convenience, we utilize a Sentence Transformer Trainer to fine-tune the embedding model. To optimize memory usage, we set the model precision to fp16 and employ Adafactor as the optimizer for model parameters. Other hyperparameters include: batch size $8$, learning rate \num{6e-6}, weight decay \num{1e-7}, max grad norm $3$.

\subsubsection{Building the IR Vector Database}

Following the fine-tuning process, we use the embedding model to convert all student responses in the training set into high-dimensional vectors, which are then used to construct a vector database. This database is indexed by these vectors, with each entry containing a JSON object that stores the original, unembedded student response and its corresponding judgment as metadata. When available, the metadata may also include the question, reference answer, and rubric.

\subsubsection{Optimizing GLM Prompt Templates}
To generate effective autoscoring prompt templates, we utilize the Claude Prompt Generator. This advanced tool transforms our initial prompt drafts into well-structured task prompts optimized for autoscoring. Our initial attempt included specified input fields, judgment criteria, and format requirements. The resulting output is a clear, step-by-step prompt encapsulated in HTML tags, optimized to guide the GLM in producing consistent and accurate autoscoring results. The full prompt template is available in Appendix \ref{apx:prompts} for reference.

\subsection{Online Running Phase}
\begin{figure}[htb]
    \centering
  \includegraphics[width=0.74\columnwidth]{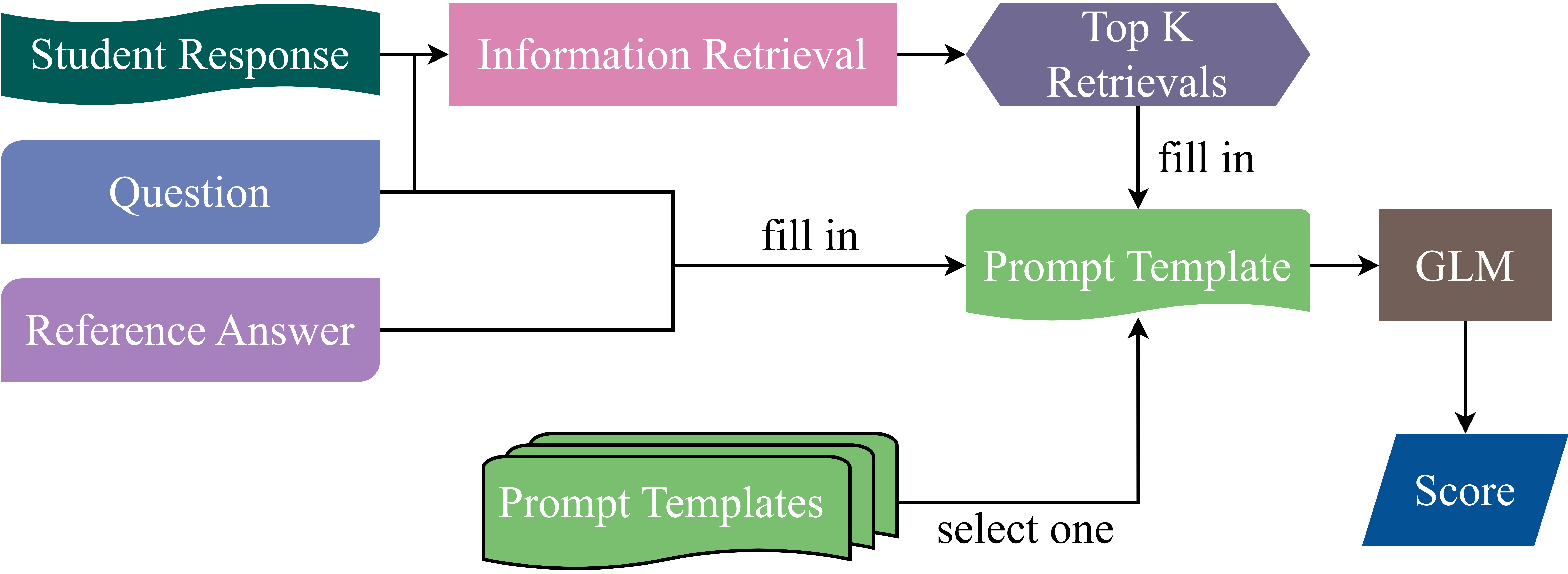}
  \caption{Proposed Approach Online Runing Workflow} \label{fig:workflow}
\end{figure}

During the online running phase, our system processes each test response through a sequence of three steps: 1) retrieve $K$ similar responses, 2) compose GLM prompt with retrievals, 3) call task GLM and parse the result. The workflow is shown in~\ref{fig:workflow}.

\subsubsection{Top-k Retrieval} 
When presented with a test response, the system first projects the response into a vector using the fine-tuned embedding model. Then, it computes cosine similarities between this vector and all vectors in the database. Subsequently, the system re-ranks all vectors based on these similarities in descending order and returns the top $k$ entries, where $k$ is a hyper-parameter that varies on the fly. This approach ensures that the most relevant historical responses are retrieved for each new test response.

\subsubsection{GLM Prompt Composition} 
The system composes the actual GLM task prompts by integrating the retrieved entries into the optimized prompt templates. While the core content of the prompts remains consistent across all tests, minor adjustments may be made to accommodate specific experimental requirements. Any modifications typically involve manually removing content that may not be applicable to certain experiments, but these adjustments are minor and do not significantly alter the prompt's overall structure or intent.

\subsubsection{GLM Autoscoring}
The final step involves sending the composed task prompts to the GLM for autoscoring. The GLM processes the prompt and generates a judgment for the student response. To ensure consistency and facilitate further analysis, the system uses regular expressions to clean and parse GLM results.
\section{Experiments}\label{sec:experiments}

\subsection{Metrics}

Many works use conventional F1 scores for evaluation. We also use them for comparison to ensure consistency with existing literature and to provide a comprehensive assessment of our model's performance. 

\begin{itemize}
\item Accuracy (ACC): This metric represents the proportion of correctly graded answers across all classes. It is calculated as: 
\[
\text{ACC} = \cfrac{\text{Number of correct predictions}}{\text{Total number of predictions}}.
\] 
Since one test sample can only be classified into one category in our task, the Accuracy is equivalent to F1 score. 

\item Macro Average F1 score (M-F1): This metric independently calculates precision, recall, and F1 scores for each label, then takes the arithmetic mean of these scores across all categories. The formula is: 
\[
\text{M-F1} = \cfrac{1}{n} \sum_{i=1}^{n} \text{ACC}_i,
\]
where $n$ is the number of categories and ACC is the Accuracy for the i-th class. This approach treats all categories equally, regardless of their size, making it particularly useful for evaluating performance on minority categories.

\item Weighted Average F1 score (W-F1): This metric calculates the F1 score for each class, then combines them by considering the proportion of test samples in each class. The formula is: 
\[
\text{W-F1} = \sum_{i=1}^{n} w_i \cdot \text{ACC}_i,
\]
where $w_i$ is the proportion of samples in the i-th class. This approach ensures that larger classes have a greater influence on the final score, reflecting their prevalence in the dataset. It provides a balance between the micro and macro averages, accounting for both class imbalance and overall performance.
\end{itemize}

\subsection{Overall Settings}

In all experiments, the proposed approach utilized a question-specific training strategy with Cosine Sentence Loss on training sets with a general labeling strategy, unless otherwise specified. All results are averaged from three runs to ensure consistency.

\subsection{Main Results}
In this section, we present the main results of our experiments across all datasets. We focus on three specific datasets: SCIENTSBANK 3-way, SCIENTSBANK 2-way, and BEETLE 5-way. It's worth noting that the SCIENTSBANK 5-way dataset is no longer publicly accessible, and therefore, it has been excluded from this study. GLM abbreviations: Haiku: Claude 3 Haiku, Sonnet: Claude 3.5 Sonnet. We compare our proposed approach with several previous state-of-the-art results. State-of-the-art results are referred from~\cite{ghavidel_using_2020}.

\subsubsection{SCIENTSBANK 3-way}

\begin{table}[htbp]
  \centering
  \caption{Main Results SCIENTSBANK 3-way.}
    \begin{tabular}{l|l|l|l|l|l|l|l|l|l}
    \toprule
    \multicolumn{1}{r}{} & \multicolumn{3}{c|}{Unseen Answer} & \multicolumn{3}{c|}{Unseen Question} & \multicolumn{3}{c}{Unseen Domain} \\
\cmidrule{2-10}    \multicolumn{1}{r}{} & Acc   & M-F1  & W-F1  & Acc   & M-F1  & W-F1  & Acc   & M-F1  & W-F1 \\
    \midrule
    COMET~\cite{ott2013comet} & 71.30 & 64.00 & 70.70 & 54.60 & 38.00 & 52.20 & 57.90 & 40.40 & 55.00 \\
    \midrule
    ETS~\cite{heilman2013ets}   & 72.00 & 64.70 & 70.80 & 58.30 & 39.30 & 53.70 & 54.30 & 33.30 & 46.10 \\
    \midrule
    SOFTCARDINALITY~\cite{jimenez2013softcardinality} & 65.90 & 55.50 & 64.70 & 65.20 & 46.90 & 63.40 & 63.70 & 48.60 & 62.00 \\
    \midrule
    Sultan et. al.~\cite{sultan2016fast} & 60.40 & 44.30 & 56.90 & 64.20 & 45.50 & 61.50 & 62.60 & 45.10 & 60.30 \\
    \midrule
    TF+SF [-question]~\cite{saha2018sentence} & 71.80 & 66.60 & 71.40 & 61.30 & 49.10 & 62.80 & 63.20 & 47.90 & 61.10 \\
    \midrule
    TF+SF [+question]~\cite{saha2018sentence} & 71.80 & 65.70 & 71.10 & 65.30 & 48.90 & 63.60 & 64.00 & 45.20 & 61.00 \\
    \midrule
    Sung's Results~\cite{sung2019improving} & 75.90 & 72.00 & 75.80 & 65.30 & 57.50 & 64.80 & 63.80 & 57.90 & 63.40 \\
    \midrule
    BERT Base uncased~\cite{ghavidel_using_2020} & 72.60 & 62.20 & 71.40 & 70.80 & 52.80 & 68.60 & 67.20 & 52.80 & 65.14 \\
    \midrule
    XLNET Base cased~\cite{ghavidel_using_2020} & 72.60 & 70.00 & 72.30 & 62.20 & 55.00 & 61.00 & 66.50 & 60.00 & 65.70 \\
    \midrule
    \midrule
    \textbf{Proposed Approach (Claude 2)} & 74.82 & \textbf{73.63} & 74.76 & \textbf{75.13} & \textbf{60.61} & \textbf{73.12} & \textbf{67.67} & 55.49 & \textbf{66.01} \\
    \midrule
    \textbf{Proposed Approach (Haiku)} & \textbf{76.52} & \textbf{74.90} & \textbf{76.44} & \textbf{73.08} & \textbf{66.31} & \textbf{72.37} & \textbf{68.84} & \textbf{65.61} & \textbf{68.49} \\
    \midrule
    \textbf{Proposed Approach (Sonnet)} & \textbf{78.71} & \textbf{78.51} & \textbf{78.64} & \textbf{76.94} & \textbf{74.43} & \textbf{76.38} & \textbf{73.35} & \textbf{73.25} & \textbf{73.04} \\
    \bottomrule
    \end{tabular}%
  \label{tab:scientsbank3way}%
\end{table}%

This experiment presents the results of the SCIENTSBANK 3-way classification task across three scenarios. The results are shown in~\ref{tab:scientsbank3way}. The Proposed Approach outperforms all other models across all scenarios and metrics. Performance generally decreases from unseen answer to unseen question and unseen domain. This limitation arises because the unseen question and unseen domain test sets contain questions that are not present in the training sets. As a result, the proposed approach cannot provide relevant examples for the RAG GLM. Specifically, by using Claude 3.5 Sonnet, our approach increases $9.04\%$, $29.44\%$, and $22.08\%$ on macro F1 score to the best previous results on unseen answer, unseen question, and unseen domain, respectively.

\subsubsection{SCIENTSBANK 2-way}

\begin{table}[htbp]
  \centering
  \caption{Main Results SCIENTSBANK 2-way.}
    \begin{tabular}{l|l|l|l|l|l|l|l|l|l}
    \toprule
    \multicolumn{1}{r}{} & \multicolumn{3}{c|}{Unseen Answer} & \multicolumn{3}{c|}{Unseen Question} & \multicolumn{3}{c}{Unseen Domain} \\
\cmidrule{2-10}    \multicolumn{1}{r}{} & Acc   & M-F1  & W-F1  & Acc   & M-F1  & W-F1  & Acc   & M-F1  & W-F1 \\
    \midrule
    COMET~\cite{ott2013comet} & 77.40 & 76.80 & 77.30 & 60.30 & 57.90 & 59.70 & 67.60 & 67.00 & 67.70 \\
    \midrule
    ETS~\cite{heilman2013ets}   & 77.60 & 76.20 & 77.00 & 63.30 & 60.20 & 62.20 & 62.70 & 54.30 & 57.40 \\
    \midrule
    SOFTCARDINALITY~\cite{jimenez2013softcardinality} & 72.40 & 71.50 & 72.20 & 74.50 & 73.70 & 74.50 & 71.10 & 70.50 & 71.20 \\
    \midrule
    Sultan et. al.~\cite{sultan2016fast} & 70.80 & 67.60 & 69.00 & 70.50 & 67.80 & 69.50 & 71.20 & 70.30 & 71.20 \\
    \midrule
    TF+SF [-question]~\cite{saha2018sentence} & 77.90 & 77.10 & 77.70 & 74.90 & 73.80 & 74.70 & 70.80 & 69.00 & 70.20 \\
    \midrule
    TF+SF [+question]~\cite{saha2018sentence} & 79.20 & 78.50 & 79.10 & 70.20 & 68.50 & 69.80 & 71.90 & 70.80 & 71.70 \\
    \midrule
    BERT Base uncased~\cite{ghavidel_using_2020} & 79.80 & 79.20 & 79.70 & 72.30 & 71.80 & 72.40 & 69.90 & 69.30 & 70.00 \\
    \midrule
    XLNET Base cased~\cite{ghavidel_using_2020} & 79.20 & 78.10 & 78.80 & 73.60 & 72.40 & 73.40 & 70.20 & 67.90 & 69.30 \\
    \midrule
    \midrule
    \textbf{Proposed Approach (Claude 2)} & \textbf{80.12} & 79.07 & \textbf{79.71} & \textbf{82.95} & \textbf{82.05} & \textbf{82.77} & \textbf{75.53} & \textbf{75.16} & \textbf{75.64} \\
    \midrule
    \textbf{Proposed Approach (Haiku)} & \textbf{81.79} & \textbf{81.57} & \textbf{81.85} & \textbf{79.67} & \textbf{78.14} & \textbf{79.18} & \textbf{75.41} & \textbf{74.74} & \textbf{75.40} \\
    \midrule
    \textbf{Proposed Approach (Sonnet)} & \textbf{82.90} & \textbf{82.12} & \textbf{82.63} & \textbf{81.17} & \textbf{79.03} & \textbf{80.23} & \textbf{76.65} & \textbf{75.11} & \textbf{76.10} \\
    \bottomrule
    \end{tabular}%
  \label{tab:scientsbank2way}%
\end{table}%

As shown in~\ref{tab:scientsbank2way}, the proposed approach also outperforms other results for the SCIENTSBANK 2-way classification task. As the 2-way task is easier than the 3-way task, our approach's performance increase is less. Specifically, by using Claude 3.5 Sonnet, our approach increases $3.69\%$, $7.09\%$, and $6.09\%$ on macro F1 score to the best previous results on unseen answer, unseen question, and unseen domain, respectively.

\subsubsection{Beetle 5-way}
We were unable to find any reliable benchmark results for the Beetle 5-way task, so we present only our own findings in Table~\ref{tab:beetle5way}. It's important to note that this is a 5-way classification task, where random guessing would yield only $20\%$ accuracy. In light of this, our proposed approach demonstrates a significant improvement, increasing the accuracy in unseen answer scenario by $267.9\%$ to reach $73.58\%$ when using Claude 3.5 Sonnet. It is interesting to find that Claude 3.5 Sonnet has a worse performance than Claude 3 Haiku in unseen question scenario.

\begin{table}[htbp]
  \centering
  \caption{Main Results BEETLE 5-way}
    \begin{tabular}{l|l|l|l|l|l|l}
    \toprule
    \multicolumn{1}{r}{} & \multicolumn{3}{c|}{Unseen Answer} & \multicolumn{3}{c}{Unseen Question} \\
\cmidrule{2-7}    \multicolumn{1}{r}{} & Acc   & M-F1  & W-F1  & Acc   & M-F1  & W-F1 \\
    \midrule
    Proposed Approach (Claude 2) & 71.75 & 62.70 & 70.87 & 23.36 & 15.65 & 20.51 \\
    \midrule
    Proposed Approach (Haiku) & 70.92 & 61.17 & 69.53 & 25.23 & 16.39 & 22.88 \\
    \midrule
    Proposed Approach (Sonnet) & 73.58 & 65.12 & 72.55 & 22.34 & 14.52 & 20.82 \\
    \bottomrule
    \end{tabular}%
  \label{tab:beetle5way}%
\end{table}%


\subsection{Ablation Study}
This section presents a series of ablation studies designed to investigate the individual contributions of key components in our proposed model. All experiments are conducted on the \textbf{SCIENTSBANK 3-way} task using the \textbf{Claude 3 Haiku model} as a trade-off between performance and cost. Through these experiments, we aim to provide insights into the relative importance of different architectural choices and training strategies.

\subsubsection{Effectiveness of input fields}
Table~\ref{tab:ablation_input_fields} shows the results of an ablation study examining the effectiveness of different input field combinations for the SCIENTSBANK 3-way task across all three scenarios. The fields Answer, Reference, and Question refer to the student's answer, reference answer, and question, respectively. The proposed approach uses all three input fields and performs best across all scenarios and metrics. In addition, the improvement is most pronounced in the unseen question and unseen domain scenarios. In detail, our approach increases $7.73\%$, $173.39\%$, and $107.24\%$ on macro F1 score on unseen answer, unseen question, and unseen domain, respectively.

\begin{table}[htbp]
  \centering
  \caption{Effectiveness of Input Fields. All results are from Claude3 Haiku.}
    \begin{tabular}{l|l|l|l|l|l|l|l|l|l}
    \toprule
    \multicolumn{1}{r}{} & \multicolumn{3}{c|}{Unseen Answer} & \multicolumn{3}{c|}{Unseen Question} & \multicolumn{3}{c}{Unseen Domain} \\
\cmidrule{2-10}    \multicolumn{1}{r}{} & Acc   & M-F1  & W-F1  & Acc   & M-F1  & W-F1  & Acc   & M-F1  & W-F1 \\
    \midrule
    Answer only & 72.08 & 69.52 & 71.88 & 46.75 & 24.26 & 34.49 & 50.79 & 31.66 & 41.91 \\
    \midrule
    Answer+Reference & 73.78 & 72.84 & 73.73 & 55.66 & 47.34 & 52.84 & 58.78 & 52.62 & 55.37 \\
    \midrule
    Answer+Question & 74.82 & 72.74 & 74.72 & 58.80 & 41.11 & 52.70 & 56.55 & 42.26 & 52.59 \\
    \midrule
    \textbf{Answer+Reference+Question} & \textbf{76.52} & \textbf{74.90} & \textbf{76.44} & \textbf{73.08} & \textbf{66.31} & \textbf{72.37} & \textbf{68.84} & \textbf{65.61} & \textbf{68.49} \\
    \bottomrule
    \end{tabular}%
  \label{tab:ablation_input_fields}%
\end{table}%



\subsubsection{Effectiveness of IR systems}

The second ablation study investigates the efficacy of various IR pipelines. The results are presented in Table~\ref{tab:ablation_ir}. "Pretrained IR" refers to the use of a pre-trained Sentence Transformer model, specifically `all-MiniLM-L6-v2,' without any fine-tuning for the IR pipeline. "Global IR" denotes the IR pipeline trained using a global training strategy, while "Question-specific IR" indicates an IR pipeline trained using a question-specific approach. The proposed approach used the question-specific training strategy with general labeling.

Results demonstrate that globally trained IR with a strict labeling strategy underperformed compared to pretrained IR. However, all other combinations outperformed the pretrained IR. Specifically, the proposed approach increased the macro-F1 score by $3.84\%$ compared to the pretrained IR.
\begin{table}[htbp]
  \centering
  \caption{Effectiveness of Fine-tuned IR. All results are from Claude3 Haiku.}
    \begin{tabular}{l|l|l|l|l}
    \toprule
    \multicolumn{1}{r}{} & \multicolumn{1}{r}{} & \multicolumn{3}{c}{Unseen Answer} \\
\cmidrule{3-5}    \multicolumn{1}{r}{} & \multicolumn{1}{r}{} & Acc   & M-F1  & W-F1 \\
\cmidrule{2-5}    \multicolumn{1}{r}{} & Pretrained IR & 73.91 & 72.13 & 73.77 \\
    \midrule
    \multirow{2}[4]{*}{Global IR} & Strict Labeling & 72.08 & 69.04 & 71.84 \\
\cmidrule{2-5}          & General Labeling & 75.12 & 72.23 & 74.98 \\
    \midrule
    \multirow{2}[4]{*}{\textbf{Question-specific IR}} & Strict Labeling & 76.09 & 74.25 & 76.00 \\
\cmidrule{2-5}          & \textbf{General Labeling} & \textbf{76.52} & \textbf{74.90} & \textbf{76.44} \\
    \bottomrule
    \end{tabular}%
  \label{tab:ablation_ir}%
\end{table}%

\subsubsection{Effectiveness of IR loss function}

Table~\ref{tab:ablation_loss_fn} presents the results of an ablation study comparing the performance of different loss functions when fine-tuning the IR pipeline. The findings indicate that the Cosine Sentence Loss demonstrated superior performance compared to the other two loss functions. Specifically, it improved the macro-F1 score by $5.92\%$ and $2.74\%$ relative to the triplet loss and cosine similarity, respectively.

\begin{table}[htbp]
  \centering
  \caption{Effectiveness of IR Loss Function. All results are from Claude3 Haiku.}
    \begin{tabular}{l|l|l|l}
    \toprule
    \multicolumn{1}{r}{} & \multicolumn{3}{c}{Unseen Answer} \\
\cmidrule{2-4}    \multicolumn{1}{r}{} & Acc   & M-F1  & W-F1 \\
    \midrule
    Triplet Loss & 74.03 & 70.71 & 73.94 \\
    \midrule
    Cosine Similarity & 75.43 & 72.90 & 75.30 \\
    \midrule
    \textbf{Cosine Sentence Loss} & \textbf{76.52} & \textbf{74.90} & \textbf{76.44} \\
    \bottomrule
    \end{tabular}%
  \label{tab:ablation_loss_fn}%
\end{table}%

\subsubsection{Effectiveness of prompts}

Table~\ref{tab:ablation_prompts} presents the results of an ablation study comparing the performance of different prompts. In particular, we compare DSPy-style prompt templates with Claude Prompt Generator-generated prompt templates. Using the Claude Prompt Generator-generated prompts, the macro-F1 score is increased by $13.14\%$, $3.15\%$, and $10.87\%$ on unseen answer, unseen question, and unseen domain compared to DSPy-style prompts respectively.

\begin{table}[htbp]
  \centering
  \caption{Effectiveness of Prompts. All results are from Claude3 Haiku.}
    \begin{tabular}{l|l|l|l|l|l|l|l|l|l}
    \toprule
    \multicolumn{1}{r}{} & \multicolumn{3}{c|}{Unseen Answer} & \multicolumn{3}{c|}{Unseen Question} & \multicolumn{3}{c}{Unseen Domain} \\
\cmidrule{2-10}    \multicolumn{1}{r}{} & Acc   & M-F1  & W-F1  & Acc   & M-F1  & W-F1  & Acc   & M-F1  & W-F1 \\
    \midrule
    DSPy-Style Prompts & 72.57 & 66.20 & 72.08 & 74.22 & 64.29 & 73.21 & 66.69 & 59.18 & 65.42 \\
    \midrule
    \textbf{CPG-Style Prompts} & \textbf{76.52} & \textbf{74.90} & \textbf{76.44} & \textbf{73.08} & \textbf{66.31} & \textbf{72.37} & \textbf{68.84} & \textbf{65.61} & \textbf{68.49} \\
    \bottomrule
    \end{tabular}%
  \label{tab:ablation_prompts}%
\end{table}%


\subsubsection{Effectiveness of RAG}

In previous experiments, all models exhibited lower F1 scores on the unseen question and unseen domain scenarios due to domain shift. To address this, we designed an experiment to explore how RAG improves autoscoring performance. The results are presented in Table~\ref{tab:ablation_rag}.

We extracted a certain fraction of the test set to build a vector database, enabling the IR pipeline to provide examples for RAG GLM autoscoring. It's important to note that the IR model was not further fine-tuned on this extracted fraction; the data was solely used for RAG purposes.

Our findings demonstrate that RAG GLM significantly enhances the macro-F1 score, with improvements of up to $13.20\%$ and $19.54\%$ in the unseen question and unseen domain scenarios, respectively. These results underscore the effectiveness of RAG in mitigating the challenges posed by domain shifts in autoscoring tasks.

\begin{table}[htbp]
  \centering
  \caption{Effectiveness of RAG. All results are from Claude3 Haiku.}
    \begin{tabular}{l|l|l|l|l|l|l}
    \toprule
    \multicolumn{1}{r}{} & \multicolumn{3}{c|}{Unseen Question} & \multicolumn{3}{c}{Unseen Domain} \\
\cmidrule{2-7}    \multicolumn{1}{r}{} & Acc   & M-F1  & W-F1  & Acc   & M-F1  & W-F1 \\
    \midrule
    UQ (20\% for VDB) & 76.28 & 72.79 & 76.19 & 78.76 & 77.30 & 78.58 \\
    \midrule
    UQ (30\% for VDB) & 76.54 & 73.15 & 76.41 & 79.71 & 78.45 & 79.59 \\
    \midrule
    \textbf{UQ (40\% for VDB)} & \textbf{77.95} & \textbf{75.79} & \textbf{77.81} & \textbf{80.26} & \textbf{78.81} & \textbf{80.16} \\
    \midrule
    UQ (50\% for VDB) & 76.66 & 73.33 & 76.52 & 79.91 & 78.28 & 79.81 \\
    \midrule
    Without RAG & 73.26 & 66.95 & 72.60 & 69.11 & 65.93 & 68.77 \\
    \bottomrule
    \end{tabular}%
  \label{tab:ablation_rag}%
\end{table}%


\section{Discussion}\label{sec:discussion}

There are many researchers who have pursued the use of GLMs for AES and ASAS \cite{chamieh2024llms, ormerod2024automated, jiang2024short}. Large proprietary GLMs present challenges for researchers due to inaccessible model weights and hardware requirements for fine-tuning. Our work showcases how RAG can effectively leverage extensive datasets without these limitations. While this approach might seem applicable to AES, we believe it is not suitable. The qualitative nature of holistic rubrics in essay evaluation contrasts with the semantic-based criteria used in short answer scoring; we believe that this might make RAG less appropriate for AES tasks.

We emphasize a key insight: RAG systems frequently rely on pretrained semantic similarity models that power these IR pipelines. Since these IR components carry out the bulk of the critical work in RAG, their accuracy can significantly impact the overall results. Our research demonstrates the substantial benefits of customizing the semantic similarity model that forms the foundation of these IR pipelines. Our ablation studies also highlight how best to fine-tune these models; using Cross-Entropy loss and question-specific similarity models. 

This pipeline can be enhanced by incorporating a follow up prompt aimed at providing feedback. In this setup, the language model could analyze the reference answer or training responses to identify key information elements necessary for a comprehensive answer. However, as with many applications of GLMs in educational settings, such a process would require rigorous oversight and validation before it could be considered for use in formative assessment programs.

\section*{Acknowledgments} \addcontentsline{toc}{section}{Acknowledgments} This paper was completed during the lead author's internship at Cambium Assessment in the summer of 2024. The authors are grateful to Kai North for his helpful suggestions.

\newpage

\bibliographystyle{unsrt}
\bibliographystyle{unsrtnat}
\bibliography{references}
\newpage
\appendix

\section{Prompt Templates} \label{apx:prompts}

\subsection{SCIENTSBANK 3-way}
\subsubsection{Unseen Answer}
This template is generated by the Claude Prompt Generator.

\begin{quote}
{\itshape
You are an expert grader tasked with evaluating answers to a given question. Your goal is to carefully analyze the provided information and make a judgment on the correctness of a new answer based on specific criteria. Follow these steps:

1. Read the following question carefully:

<question>

\{\{QUESTION\}\}

</question>

2. Now, consider the reference answer, which is the gold standard for correctness:

<reference\_answer>

\{\{REFERENCE\_ANSWER\}\}

</reference\_answer>

3. Review the following examples of similar answers and their corresponding judgments from a golden grader:

<examples>

\{\{EXAMPLES\}\}

</examples>

4. Now, examine the new answer that needs to be evaluated:

<new\_answer>

\{\{NEW\_ANSWER\}\}

</new\_answer>

5. Analyze the new answer by comparing it to the reference answer and the examples provided. Consider the following criteria:

   a. Is the new answer a complete paraphrase of the reference answer that correctly answers the question?
   
   b. Is the new answer partially correct, irrelevant to the question, or out of the domain?
   
   c. Does the new answer explicitly contradict the reference answer?

6. Based on your analysis, choose one of the following judgments:

   - "correct": if the new answer is a complete paraphrase of the reference answer that answers the question correctly.
   
   - "incorrect": if the new answer is partially correct, irrelevant to the question, or out of the domain.
   
   - "contradictory": if the new answer explicitly contradicts the reference answer.

7. Provide your judgment using the following format:

<judgment>

[Insert your chosen judgment here: correct, incorrect, or contradictory]

</judgment>

Important notes:

- Do not provide any explanations or additional text apart from the judgment itself.

- Ensure that your judgment aligns with the most similar answer in the examples provided.

- Focus solely on the content and meaning of the answers, not on minor differences in wording or style.}
\end{quote}

\subsubsection{Unseen Question \& Unseen Domain}
This template is derived from the above template by removing \textit{Examples} related content.

\begin{quote}
{\itshape
You are an expert grader tasked with evaluating answers to a given question. Your goal is to carefully analyze the provided information and make a judgment on the correctness of a new answer based on specific criteria. Follow these steps:

1. Read the following question carefully:

<question>

\{\{QUESTION\}\}

</question>

2. Now, consider the reference answer, which is the gold standard for correctness:

<reference\_answer>

\{\{REFERENCE\_ANSWER\}\}

</reference\_answer>

3. Now, examine the new answer that needs to be evaluated:

<new\_answer>

\{\{NEW\_ANSWER\}\}

</new\_answer>

4. Analyze the new answer by comparing it to the reference answer provided. Consider the following criteria:

   a. Is the new answer a complete paraphrase of the reference answer that correctly answers the question?
   
   b. Is the new answer partially correct, irrelevant to the question, or out of the domain?
   
   c. Does the new answer explicitly contradict the reference answer?

5. Based on your analysis, choose one of the following judgments:

   - "correct": if the new answer is a complete paraphrase of the reference answer that answers the question correctly.
   
   - "incorrect": if the new answer is partially correct, irrelevant to the question, or out of the domain.
   
   - "contradictory": if the new answer explicitly contradicts the reference answer.

6. Provide your judgment using the following format:

<judgment>

[Insert your chosen judgment here: correct, incorrect, or contradictory]

</judgment>

Important notes:

- Do not provide any explanations or additional text apart from the judgment itself.

- Focus solely on the content and meaning of the answers, not on minor differences in wording or style.}
\end{quote}

\subsubsection{Unseen Answer}
This template is optimized by DSPy.

\begin{quote}
{\itshape
You are an expert grader. Read the question, reference answer and examples carefully, learn how the golden grader evaluated answers in examples, and then grade the new answer with the same criteria. Choice of Judgment: (A) correct: if the new answer is a complete and correct paraphrase of the reference answer. (B) incorrect: if the new answer is partially correct, irrelevant, or out of the domain. (C) contradictory: if the new answer explicitly contradicts the reference answer.

---

Follow the following format.

Question: the question that all answers targeted.

Reference Answer: the gold answer for reference.

Examples: similar answers and corresponding judgments from a golden grader.

New Answer: should have the same judgment with its most similar answer in the example.

Judgment of the New Answer: choose from {`correct`, `incorrect`, `contradictory`}. Do not provide any explanations or text apart from the judgment.

---

Question: \{\{QUESTION\}\}

Reference Answer: \{\{REFERENCE\_ANSWER\}\}

Examples:

\{\{EXAMPLES\}\}

New Answer: \{\{NEW\_ANSWER\}\}

Judgment of the New Answer:
}
\end{quote}

\subsubsection{Unseen Question \& Unseen Domain}
This template is derived from the above template by removing \textit{Examples} related content.

\begin{quote}
{\itshape
You are an expert grader. Read the question and reference answer carefully then grade the new answer. Choice of Judgment: (A) correct: if the new answer is a complete and correct paraphrase of the reference answer. (B) incorrect: if the new answer is partially correct, irrelevant, or out of the domain. (C) contradictory: if the new answer explicitly contradicts the reference answer.

---

Follow the following format.

Question: the question that all answers targeted.

Reference Answer: the gold answer for reference.

New Answer: should have the same judgment with its most similar answer in the example.

Judgment of the New Answer: choose from {`correct`, `incorrect`, `contradictory`}. Do not provide any explanations or text apart from the judgment.

---

Question: \{\{QUESTION\}\}

Reference Answer: \{\{REFERENCE\_ANSWER\}\}

New Answer: \{\{NEW\_ANSWER\}\}

Judgment of the New Answer:
}
\end{quote}

\subsection{SCIENTSBANK 2-way}
\subsubsection{Unseen Answer}
This template is generated by the Claude Prompt Generator.

\begin{quote}
{\itshape
You are an expert grader tasked with evaluating answers to a given question. Your goal is to carefully analyze the provided information and make a judgment on the correctness of a new answer based on specific criteria. Follow these steps:

1. Read the following question carefully:

<question>

\{\{QUESTION\}\}

</question>

2. Now, consider the reference answer, which is the gold standard for correctness:

<reference\_answer>

\{\{REFERENCE\_ANSWER\}\}

</reference\_answer>

3. Review the following examples of similar answers and their corresponding judgments from a golden grader:

<examples>

\{\{EXAMPLES\}\}

</examples>

4. Now, examine the new answer that needs to be evaluated:

<new\_answer>

\{\{NEW\_ANSWER\}\}

</new\_answer>

5. Analyze the new answer by comparing it to the reference answer and the examples provided. Consider the following criteria: is the new answer a complete paraphrase of the reference answer that correctly answers the question?

6. Based on your analysis, choose one of the following judgments:

   - "correct": if the new answer is a complete paraphrase of the reference answer that answers the question correctly.
   
   - "incorrect": if the new answer is partially correct, contradictory or irrelevant to the question, or out of the domain.

7. Provide your judgment using the following format:

<judgment>

[Insert your chosen judgment here: correct, incorrect, or contradictory]

</judgment>

Important notes:

- Do not provide any explanations or additional text apart from the judgment itself.

- Ensure that your judgment aligns with the most similar answer in the examples provided.

- Focus solely on the content and meaning of the answers, not on minor differences in wording or style.
}
\end{quote}

\subsubsection{Unseen Question \& Unseen Domain}
This template is derived from the above template by removing \textit{Examples} related content.

\begin{quote}
{\itshape
You are an expert grader tasked with evaluating answers to a given question. Your goal is to carefully analyze the provided information and make a judgment on the correctness of a new answer based on specific criteria. Follow these steps:

1. Read the following question carefully:

<question>

\{\{QUESTION\}\}

</question>

2. Consider the reference answer, which is the gold standard for correctness:

<reference\_answer>

\{\{REFERENCE\_ANSWER\}\}

</reference\_answer>

3. Now, examine the new answer that needs to be evaluated:

<new\_answer>

\{\{NEW\_ANSWER\}\}

</new\_answer>

4. Analyze the new answer by comparing it to the reference answer. Consider the following criteria: is the new answer a complete paraphrase of the reference answer that correctly answers the question?

5. Based on your analysis, choose one of the following judgments:

   - "correct": if the new answer is a complete paraphrase of the reference answer that answers the question correctly.
   
   - "incorrect": if the new answer is partially correct, contradictory or irrelevant to the question, or out of the domain.

6. Provide your judgment using the following format:

<judgment>

[Insert your chosen judgment here: correct, incorrect, or contradictory]

</judgment>

Important notes:

- Do not provide any explanations or additional text apart from the judgment itself.

- Focus solely on the content and meaning of the answers, not on minor differences in wording or style.
}
\end{quote}

\subsection{BEETLE 5-way}
\subsubsection{Unseen Answer}

\begin{quote}
{\itshape
You are an expert grader tasked with evaluating answers to a given question. Your goal is to carefully analyze the provided information and make a judgment on the correctness of a new answer based on specific criteria. Follow these steps:

1. Read the following question carefully:

<question>

\{\{QUESTION\}\}

</question>

2. Now, consider the reference answer, which is the gold standard for correctness:

<reference\_answer>

\{\{REFERENCE\_ANSWER\}\}

</reference\_answer>

3. Review the following examples of similar answers and their corresponding judgments from a golden grader:

<examples>

\{\{EXAMPLES\}\}

</examples>

4. Now, examine the new answer that needs to be evaluated:

<new\_answer>

\{\{NEW\_ANSWER\}\}

</new\_answer>

5. Analyze the new answer by comparing it to the reference answer and the examples provided. Consider the following criteria:

   - Completeness: Does the answer fully address all aspects of the question?
   
   - Accuracy: Is the information provided correct and consistent with the reference answer?
   
   - Relevance: Does the answer directly relate to the question asked?
   
   - Contradictions: Are there any statements that contradict the reference answer?
   
   - Domain knowledge: Does the answer demonstrate appropriate understanding of the subject matter?

6. Based on your analysis, choose one of the following judgments:

   - "correct": when the student answer is a complete and correct representation of one of the reference answers.
   
   - "partially correct but incomplete": when the student answer is correct to the extent it is written, but is not complete.
   
   - "contradictory": when the student answer contradicts the reference answer, i.e., both cannot be correct at the same time.
   
   - "irrelevant": when the student answer is irrelevant to the reference answer although it may still be talking about the reference answer.
   
   - "non-domain": when the student answer lacks domain content, i.e., the student may be asking for 'help', 'advice' etc. like 'I don't know', 'what the book says', 'you are stupid'.

7. Provide your judgment using the following format:

<judgment>

[Insert your chosen judgment here: correct, partially correct but incomplete, contradictory, irrelevant, or non-domain]

</judgment>

Important notes:

- Do not provide any explanations or additional text apart from the judgment itself.

- Ensure that your judgment aligns with the most similar answer in the examples provided.

- Focus solely on the content and meaning of the answers, not on minor differences in wording or style.
}
\end{quote}

\subsubsection{Unseen Question \& Unseen Domain}
This template is derived from the above template by removing \textit{Examples} related content.

\begin{quote}
{\itshape
You are an expert grader tasked with evaluating answers to a given question. Your goal is to carefully analyze the provided information and make a judgment on the correctness of a new answer based on specific criteria. Follow these steps:

1. Read the following question carefully:

<question>

\{\{QUESTION\}\}

</question>

2. Consider the reference answer, which is the gold standard for correctness:

<reference\_answer>

\{\{REFERENCE\_ANSWER\}\}

</reference\_answer>

3. Now, examine the new answer that needs to be evaluated:

<new\_answer>

\{\{NEW\_ANSWER\}\}

</new\_answer>

4. Analyze the new answer by comparing it to the reference answer provided. Consider the following criteria:

   - Completeness: Does the answer fully address all aspects of the question?
   
   - Accuracy: Is the information provided correct and consistent with the reference answer?
   
   - Relevance: Does the answer directly relate to the question asked?
   
   - Contradictions: Are there any statements that contradict the reference answer?
   
   - Domain knowledge: Does the answer demonstrate appropriate understanding of the subject matter?

5. Based on your analysis, choose one of the following judgments:

   - "correct": when the student answer is a complete and correct representation of one of the reference answers.
   
   - "partially correct but incomplete": when the student answer is correct to the extent it is written, but is not complete.
   
   - "contradictory": when the student answer contradicts the reference answer, i.e., both cannot be correct at the same time.
   
   - "irrelevant": when the student answer is irrelevant to the reference answer although it may still be talking about the reference answer.
   
   - "non-domain": when the student answer lacks domain content, i.e., the student may be asking for 'help', 'advice' etc. like 'I don't know', 'what the book says', 'you are stupid'.

6. Provide your judgment using the following format:

<judgment>

[Insert your chosen judgment here: correct, partially correct but incomplete, contradictory, irrelevant, or non-domain]

</judgment>

Important notes:

- Do not provide any explanations or additional text apart from the judgment itself.

- Focus solely on the content and meaning of the answers, not on minor differences in wording or style.
}
\end{quote}
\end{document}